# SPATIALLY DIRECTIONAL PREDICTIVE CODING FOR BLOCK-BASED COMPRESSIVE SENSING OF NATURAL IMAGES

*Jian Zhang, Debin Zhao, Feng Jiang*

School of Computer Science and Technology, Harbin Institute of Technology (HIT), China

**ABSTRACT**

A novel coding strategy for block-based compressive sensing named spatially directional predictive coding (SDPC) is proposed, which efficiently utilizes the intrinsic spatial correlation of natural images. At the encoder, for each block of compressive sensing (CS) measurements, the optimal prediction is selected from a set of prediction candidates that are generated by four designed directional predictive modes. Then, the resulting residual is processed by scalar quantization (SQ). At the decoder, the same prediction is added onto the de-quantized residuals to produce the quantized CS measurements, which is exploited for CS reconstruction. Experimental results substantiate significant improvements achieved by SDPC-plus-SQ in rate distortion performance as compared with SQ alone and DPCM-plus-SQ.

*Index Terms*— Compressive sensing, directional prediction, predictive coding, scalar quantization

## 1. INTRODUCTION

Recent years have witnessed the rapid development of Compressive Sensing (CS), which has drawn quite an amount of attention in the society of signal processing [1-12]. From many fewer acquired measurements than suggested by the Nyquist sampling theory, CS theory demonstrates that a signal can be reconstructed with high probability when it exhibits sparsity in some domain, which has greatly changed the way engineers think of data acquisition.

Most CS literature assumes that the measurement process is effectuated within the hardware of the sensing device, where lower-dimensional CS measurements are obtained with respect to original high-dimensional data. Accordingly, CS measurement process can be regarded as conducting data acquisition and data compression simultaneously. However, the data compression CS measurement process brings is not a real compression in the strict information-theoretic sense, but a type of dimensionality reduction in essence [9]. Real compression is the process that produces an ultimate compressed bitstream from the input data, which is done by the so called encoder. Therefore, some form of quantization is desirable to generate a compressed bitstream from the CS measurements to implement the real CS compression, which has been largely omitted by most CS literature.

One straightforward solution is to simply apply scalar quantization (SQ) to CS measurements acquired by the sensing device. Nonetheless, due to the ignorance of characteristics of CS and signal itself, the simple solution is found highly inefficient in rate-distortion performance [8, 9]. As a result, much attention in recent literature has been devoted to the improvement of rate-distortion performance of quantized CS, mainly depending on the quantization optimization process [7], the reconstruction process [5], or both [4], which appears very complicated.

Lately, as opposed to the works mentioned before, *Mun* and *Fowler* [9] proposed a framework of quantization via simple uniform SQ coupled with differential pulse code modulation (DPCM) of the CS measurements, which is applicable to the CS of images effectuated in blocks, i.e., block-based CS (BCS) [2, 3, 10, 12]. Particularly, the previous block is considered as a prediction and subtracted from the current block of measurements in the measurement domain. Instead of applying quantization directly to each block of CS measurements, the resulting residual is then scalar-quantized. The simple DPCM-plus-SQ approach to quantized CS provides surprisingly competitive rate-distortion performance. However, it is not efficient enough to leverage previous block as the prediction of current one, since the non-stationarity of natural images is ignored.

Inspired by intra prediction in video coding, this paper greatly extends the previous work [9] and proposes a novel coding strategy for block-based compressive sensing, called spatially directional predictive coding (SDPC), which efficiently utilizes the intrinsic spatial correlation of natural images. For each block of compressive sensing measurements, its optimal prediction is selected from a set of multiple prediction candidates that are generated by four designed directional predictive modes. To the best of our knowledge, this is the first time that directional predictive coding is incorporated into the framework of block compressive sensing. Experimental results verify the effectiveness of our proposed SDPC.

This paper is organized as follows. Section 2 briefly reviews CS and BCS. The details of proposed SDPC are provided Section 3. Experimental results are shown in Section 4, and conclusions are drawn in Section 5.

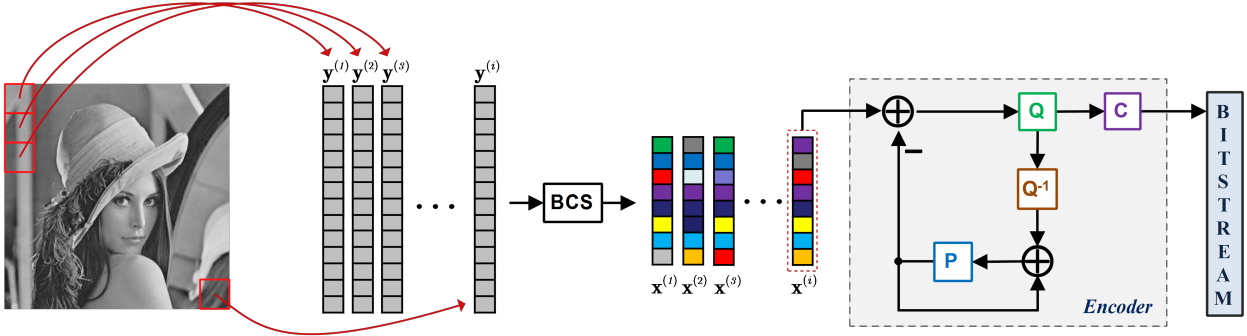

Figure 1: Architecture of SDPC-plus-SQ to block-based compressive sensing (BCS). BCS is implemented with any CS-based image-acquisition; Q is uniform SQ; $Q^{-1}$ is inverse uniform SQ; C is entropy coder; P is proposed spatially directional predictive coding module.

## 2. BACKGROUND

Briefly, if a signal is sparse in frequency domain, or in some incoherent domain by extension, CS allows exact recovery of the signal from its time or space measurements acquired by linear projection, whose number is much smaller than that of the original signal. More specifically, suppose that we have a signal $\mathbf{y} \in \Re^N$ and its measurements $\mathbf{x} \in \Re^M$, namely, $\mathbf{x} = \Phi \mathbf{y}$. Here, $\Phi$ is an $M \times N$ measurement matrix such that $M$ is much smaller than $N$. Our purpose is to recover $\mathbf{y}$ from $\mathbf{x}$ with subrate, being $S = M/N$.

In order to avoid the large storage cost of measurement matrix, an alternative paradigm for the CS of 2D images was proposed [2], wherein the sampling of an image is driven by random matrices applied on a block-by-block basis, i.e., block-based CS (BCS). That is, an image $\mathbf{y}$ is first divided into $n$ non-overlapped blocks of size $B \times B$ with each block denoted by $\mathbf{y}^{(i)} \in \Re^{B^2}, i = 1, 2, ..., n$ in vector representation along the vertical or horizontal scan order. Then, its corresponding measurements $\mathbf{x}^{(i)}$ is obtained by

$$\mathbf{x}^{(i)} = \Phi_B \mathbf{y}^{(i)}, \quad (1)$$

where $\mathbf{x}^{(i)} \in \Re^{M_B}$ and $\Phi_B$ is an $M_B \times B^2$ measurement matrix such that the subrate for the image as a whole remains $S = M_B/B^2$. It is straightforward to conclude that $\Phi_B$ applied to an image at block level is equivalent to a whole image measurement matrix $\Phi$ with a constrained structure, namely, $\Phi$ can be written as a block diagonal with $\Phi_B$ along the diagonal [3].

## 3. THE PROPOSED SPATIALLY DIRECTIONAL PREDICTIVE CODING (SDPC)

Since each block of a natural image is not isolated, which has intrinsic spatial correlation with its neighboring blocks; it is straightforward to expect that each block of CS measurements are also highly related with those of its neighboring blocks. Further, because different blocks may have different directional correlations due to the non-stationarity of natural images, the optimal directional mode for each block should be selected from multiple candidates. This is just the tenet of our proposed spatially directional predictive coding (SDPC) for block-based CS. The details of SDPC are given below.

As shown in Fig. 1, an input image $\mathbf{y}$ is first divided into $n$ non-overlapped blocks of size $B \times B$ with each block denoted by $\mathbf{y}^{(i)} \in \Re^{B^2}, i = 1, 2, ..., n$ in vector form along the horizontal scan order. Then, all blocks of CS measurements are acquired by Eq. (1). Next, borrowing the idea of intra prediction in H.264 for 16×16 macroblock, for $i$th block of measurements, denoted by $\mathbf{x}^{(i)}$, we design four directional prediction modes from its neighboring already reconstructed measurements, namely, vertical, horizontal, DC, and diagonal, as illustrated in Fig. 2. It is important to stress that the main difference between intra prediction in video coding and our directional prediction for CS lies that the former is at the level of pixel, in space domain while the latter is at the level of block in CS measurement domain. More specially, let $\tilde{\mathbf{x}}_A^{(i)}$, $\tilde{\mathbf{x}}_B^{(i)}$, and $\tilde{\mathbf{x}}_C^{(i)}$ denote the up-left, up, and left blocks of measurements with regard to $\mathbf{x}^{(i)}$, respectively. As illustrated in Fig. 2, the corresponding predictions by four modes above are defined:

*Vertical Mode:*
$$\hat{\mathbf{x}}_V^{(i)} = \tilde{\mathbf{x}}_B^{(i)}, \quad (2)$$

*Horizontal Mode:*
$$\hat{\mathbf{x}}_H^{(i)} = \tilde{\mathbf{x}}_C^{(i)}, \quad (3)$$

*DC Mode:*
$$\hat{\mathbf{x}}_{DC}^{(i)} = (\tilde{\mathbf{x}}_B^{(i)} + \tilde{\mathbf{x}}_C^{(i)}) \gg 1, \quad (4)$$

*Diagonal Mode:*
$$\hat{\mathbf{x}}_{Diag}^{(i)} = \tilde{\mathbf{x}}_A^{(i)}, \quad (5)$$

where the symbol $\gg$ denotes the right shift operator. It is worth noting that, in [9] each block is also extracted in a vertical scan order. Thus, it is obvious to witness that DPCM [9], which exploits previous block as prediction, is equivalent to proposed vertical mode, just a special case of our proposed SDPC. Accordingly, it is believed that SDPC can gain higher performance than DPCM.

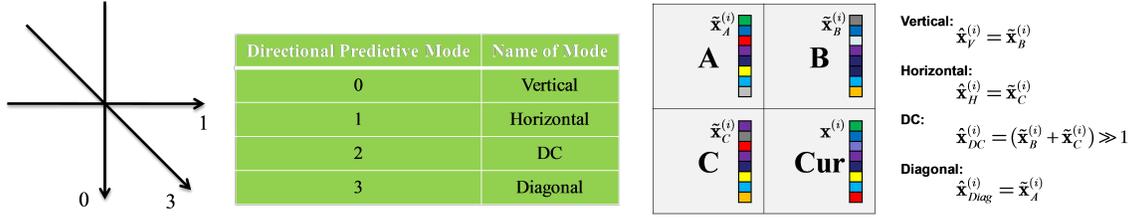

**Figure 2:** Illustrations of proposed four directional predictive modes and four corresponding predictive results with respect to current block of measurements $\mathbf{x}^{(i)}$ from its neighboring already reconstructed measurements.

Now we define the set $\Xi$, $\Xi = \{\hat{\mathbf{x}}_V^{(i)}, \hat{\mathbf{x}}_H^{(i)}, \hat{\mathbf{x}}_{DC}^{(i)}, \hat{\mathbf{x}}_{Diag}^{(i)}\}$, which is a collection of four directional prediction results above. The optimal prediction, denoted by $\hat{\mathbf{x}}_P^{(i)}$, for the measurements of the current block is then determined by minimizing the residual between $\mathbf{x}^{(i)}$ and the measurement of four predictive results in $\Xi$, i.e.,

$$\hat{\mathbf{x}}_P^{(i)} = \mathrm{argmin}_{\mathbf{x} \in \Xi} \| \mathbf{x} - \mathbf{x}^{(i)} \|_{\ell_1}. \quad (6)$$

Here, $\| * \|_{\ell_1}$ is $\ell_1$ norm, adding all the absolute values of the entries in a vector.

After obtaining the optimal prediction of $\mathbf{x}^{(i)}$, the residual can be calculated by $\mathbf{d}^{(i)} = \mathbf{x}^{(i)} - \hat{\mathbf{x}}_P^{(i)}$, which is then scalar-quantized to acquire the quantization index $\mathbf{s}^{(i)} = \mathrm{Q}[\mathbf{d}^{(i)}]$. The operation of de-quantization of $\mathbf{s}^{(i)}$ is then conducted to get the quantized residual $\tilde{\mathbf{d}}^{(i)}$, which is then added by $\hat{\mathbf{x}}_P^{(i)}$, producing the reconstructed CS measurements of $\mathbf{x}^{(i)}$, denoted by $\tilde{\mathbf{x}}^{(i)}$, ready for further prediction coding.

What is needed to be written into bitstream is composed of two parts: the flag of best predictive mode (2 bit) and the bits to encode $\mathbf{s}^{(i)}$ by entropy coder. Note that 2 bit overhead for each block is almost neglectable as compared with the exciting gains it brings (for instance, if block size is set to be 16×16, then the overhead is only 2/256 = 0.0078 bpp). This process is applied for all blocks of CS measurements to achieve the final bitstream.

In the decoder side, similar to the procedures before, all block of reconstructed measurements are obtained from the bitstream, which are then utilized for ultimate image reconstruction by image CS recovery algorithms.

To demonstrate the superiority of SDPC over DPCM, we make a further quantitative comparison. Like [9], we also use the correlation coefficient to measure the correlation of two blocks to measurements $\mathbf{x}^{(i)}, \mathbf{x}^{(j)}$, defined as

$$\rho(\mathbf{x}^{(i)}, \mathbf{x}^{(j)}) = \frac{\mathbf{x}^{(i)T}\mathbf{x}^{(j)}}{\|\mathbf{x}^{(i)}\|\|\mathbf{x}^{(j)}\|}. \quad (7)$$

For each block, we calculate its two types of correlation coefficients, i.e., $CC_1^{(i)}$ and $CC_2^{(i)}$, which are defined as $CC_1^{(i)} = \rho(\mathbf{x}^{(i)}, \mathbf{x}^{(i-1)})$ and $CC_2^{(i)} = \rho(\mathbf{x}^{(i)}, \mathbf{x}_P^{(i)})$, where $\mathbf{x}^{(i-1)}$ is the measurements of previous block of $\mathbf{x}^{(i-1)}$ and $\mathbf{x}_P^{(i)}$ is its optimal prediction selected by our SDPC from four candidates.

Table I provides the corresponding two types of average correlation coefficients (ACC), which are computed from the above $CC_1^{(i)}$ and $CC_2^{(i)}$ over all blocks for three test images wherein 16×16 blocks are extracted from images and subject to random projection with a subrate of 0.5. There is no doubt that $ACC_2$ is larger than $ACC_1$ for all images, which indicates that the prediction by SDPC is more efficient than that by DPCM. Additionally, the percentage of each predictive mode in SDPC to compute $ACC_2$ is presented in Table II, which clearly illustrates the contributions of each predictive mode.

Table I: Average Correlation Coefficient in Measurement-Domain

| Image | Clown | Peppers | Lenna | Avg. |
|---|---|---|---|---|
| $ACC_1$ | 0.8630 | 0.9509 | 0.9713 | 0.9284 |
| $ACC_2$ | 0.9043 | 0.9679 | 0.9777 | 0.9500 |

Table II: Percentage of Each Predictive Mode in SDPC to Compute $ACC_2$

| Mode | Clown | Peppers | Lenna | Avg. |
|---|---|---|---|---|
| 0 | 39.75% | 39.55% | 50.00% | 43.10% |
| 1 | 21.88% | 27.18% | 12.79% | 20.62% |
| 2 | 28.32% | 30.08% | 27.34% | 28.58% |
| 3 | 9.96% | 8.50% | 9.77% | 4.41% |

## 4. EXPERIMENTAL RESULTS

Experimental results are provided to verify the performance of the proposed technique SDPC for block-based CS compressive sensing of natural images. The rate-distortion efficiency of SDPC-plus-SQ is examined by comparing it to DPCM-plus-SQ and SQ applied alone to BCS measurements. Two image CS recovery algorithms, namely, SPL [3] and CoS [6], are exploited to effectuate CS recovery from the decoded measurements generated by the above three comparative techniques. The implementations of SPL[1], DPCM[1], CoS[2] as well as SDPC[3] can be found at the corresponding websites.

---

[1] http://www.ece.msstate.edu/~fowler/BCSSPL/.
[2] http://idm.pku.edu.cn/staff/zhangjian/RCoS/.
[3] http://idm.pku.edu.cn/staff/zhangjian/SDPC/.

Table III: PSNR (in dB) Performance Comparison for Various Bitrates

| Image | Clown | | | Peppers | | | Lenna | | | Avg. |
|---|---|---|---|---|---|---|---|---|---|---|
| Bitrate (bpp) | 0.3 | 0.4 | 0.5 | 0.3 | 0.4 | 0.5 | 0.3 | 0.4 | 0.5 | |
| SQ+SPL | 25.8 | 26.5 | 27.2 | 27.2 | 28.0 | 28.6 | 27.1 | 27.8 | 28.4 | 27.4 |
| DPCM-plus-SQ+SPL | 26.5 | 27.6 | 28.2 | 28.1 | 29.2 | 30.0 | 28.6 | 29.4 | 30.0 | 28.6 |
| SDPC-plus-SQ+SPL | 27.4 | 28.2 | 29.0 | 29.0 | 29.9 | 30.6 | 29.0 | 29.8 | 30.4 | 29.3 |
| *Gain* | +0.9 | +0.6 | +0.8 | +0.9 | +0.7 | +0.6 | +0.4 | +0.4 | +0.4 | +0.6 |
| SQ+CoS | 27.5 | 28.6 | 29.6 | 28.7 | 29.3 | 30.1 | 27.9 | 28.6 | 29.4 | 28.9 |
| DPCM-plus-SQ+CoS | 27.9 | 29.5 | 30.8 | 29.4 | 30.8 | 31.6 | 29.6 | 30.6 | 31.3 | 30.2 |
| SDPC-plus-SQ+CoS | 29.1 | 30.6 | 31.8 | 30.4 | 31.4 | 32.1 | 30.0 | 30.9 | 31.6 | 30.9 |
| *Gain* | +1.2 | +1.1 | +1.0 | +1.0 | +0.6 | +0.5 | +0.4 | +0.3 | +0.3 | +0.7 |

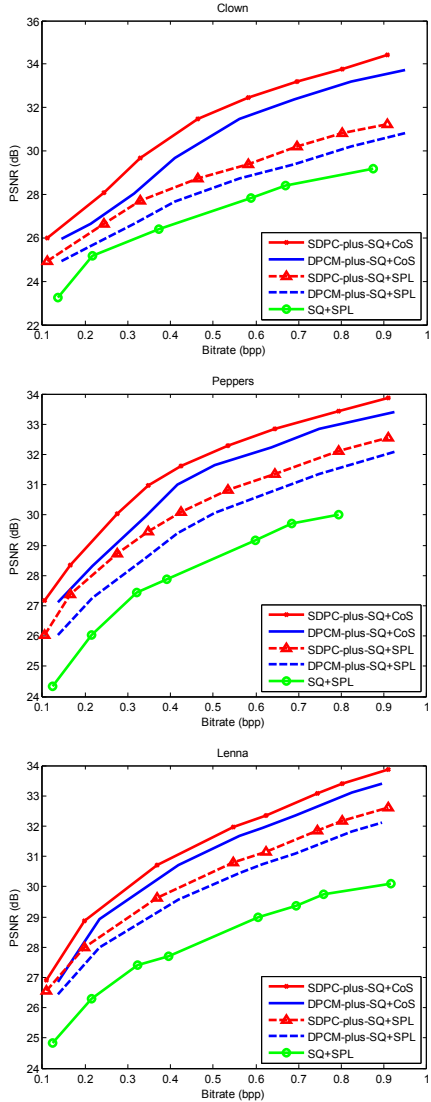

Figure 3: Rate-distortion performance for various images from 0.1 bpp to 1.0 bpp.

In our implementations, the three grayscale test images are of size 512×512, the block size for BCS is set to be 16×16 and the measurement matrix $\Phi_B$ is an orthogonal random Gaussian matrix. The rate-distortion performance in terms of peak signal-to-noise ratio (PSNR) in dB and bitrate in bits per pixel (bpp) is provided. Following [9], the actual bitrate is estimated using the entropy of the quantizer indices, which would be actually produced by a real entropy coder. The setup of the combination of quantizer step-size and subrate is the same as that in [9].

Table III gives the PSNR results at various bitrates (from 0.3 bpp to 0.5 bpp) for the two BCS recovery algorithms, i.e., SPL and CoS, coupled with three coding techniques, namely, SQ alone, DPCM-plus-SQ and SDPC-plus-SQ. Note that the results of DPCM-plus-SQ in Table III differ slightly from those provided in [9], since different random projections are used. One easily observes that, for both SPL and CoS, SDPC-plus-SQ improves 2.0 dB and 1.2 dB gain on average PSNR as compared to SQ alone and DPCM-plus-SQ, respectively. In addition, Figs. 3 presents the rate-distortion performance for different BCS recovery algorithms combined with different BCS coding techniques for a bitrate ranging from 0.1 to 1.0 bpp, which sufficiently demonstrates the superiority of SDPC over DPCM.

## 5. CONCLUSION

A novel coding strategy for block-based compressive sensing, named spatial directional predictive coding (SDPC), is proposed, which efficiently utilizes the intrinsic spatial correlation of natural images. To the best of our knowledge, this is the first time that directional predictive coding is incorporated into the framework of block compressive sensing. Experimental results substantiate significant improvements by SDPC-plus-SQ as compared with SQ alone and DPCM-plus-SQ.

## 6. ACKNOWLEDGEMENT

This work was supported in part by National Basic Research Program of China (973 Program 2009CB320905), Natural Science Foundation of Jiangsu Province under Grant BK2012397 and by the National Science Foundation of China under Grants 61272386 and 61100096.